# MHfit: Mobile Health Data for Predicting Athletics Fitness Using Machine Learning

Jonayet Miah
*Department of Computer Science*
*University of South Dakota*
South Dakota, USA
jonayet.miah@coyotes.usd.edu

Muntasir Mamun
*Department of Computer Science*
*University of South Dakota*
South Dakota, USA
muntasir.mamun@coyotes.usd.edu

Md Minhazur Rahman
*Department of Computer Science*
*University of South Dakota*
South Dakota, USA
minhazur.rahman@coyotes.usd.edu

Md Ishtyaq Mahmud
*College of Science & Engineering*
*Central Michigan University*
Mount Pleasant, MI 48858, USA
mahmu4m@cmich.edu

Sabbir Ahmad
*Department of Mathematical Science*
*University of South Dakota*
South Dakota, USA
sabbir.ahmad@coyotes.usd.edu

Md Hasan Bin Nasir
*Department of Computer Science*
*University of South Dakota*
South Dakota, USA
mdhasanbin.nasir@coyotes.usd.edu

*Abstract*— Mobile phones and other electronic gadgets/devices have aided in collecting data without the need for data entry. This paper will specifically focus on Mobile health data(m-health). Mobile health data use mobile devices to gather clinical health data and track patients' vitals in real-time. Our study is aimed to give decisions for small or big sports teams on whether one athlete good fit or not for a particular game with the compare several machine learning algorithms to predict human behavior and health using the data collected from mobile devices and sensors placed on patients. In this study, we have obtained the dataset from a similar study done on m-health. The dataset contains vital signs recordings of ten volunteers from different backgrounds. They had to perform several physical activities with a sensor placed on their bodies. Our study used 5 machine learning algorithms (XGBoost, Naïve Bayes, Decision Tree, Random Forest, and Logistic Regression) to analyze and predict human health behavior. XGBoost performed better compared to the other machine learning algorithms and achieved 95.2% in accuracy, 99.5% in sensitivity, 99.5% in specificity, and 99.66% in F-1 score. Our research indicated a promising future in m-health being used to predict human behavior and further research and exploration need to be done for it to be available for commercial use specifically in the sports industry.

## I. Introduction

Personal data is being regarded as a new economic asset. Our smartphone's personal data can be utilized for a variety of purposes, including identification, recommendation systems, predicting personalities by analyzing patterns in human behavior, and logging human health data with sensors. Our research will focus on mobile health for human behavior analysis. Artificial intelligence (AI) gives us insight into data that can be used to revolutionize and transform different industries that will propel mankind to new heights. Artificial intelligence, specifically in the healthcare industry, can have a major impact on saving lives by providing solutions to pressing issues seen in healthcare. Mobile health is considered one of the main drivers that are seeking to bring this transformative change. According to M-health: Fundamentals and

Applications book, Mobile Health was first defined as 'mobile computing, medical sensors, and communication technologies for healthcare [1]. The technological breakthrough connected with the release of the first smartphone had a significant impact on the evolution of m-Health. This also allowed for the creation of powerful embedded computational tools in smartphones and other technological devices(e.g., smart watches, wearable monitors, and sensors) to produce massive volumes of mobile health data. This innovative move ushered in the smartphonecentric m-Health age along with beginning of the smartphonefocused m-Health era. The patient-centered approach is starting to take center stage in the paradigm. Smartphone M-Health is becoming a reality. We can get real-time data from apps by connecting them to a variety of wearable sensors. Recent developments have led to the emergence of m-health 2.0, which is described as "the convergence of m-health with emerging development in smart sensors, 5G communications systems with the functional capabilities of web 2.0, cloud computing, and social networking technologies, toward personalized patient-centered healthcare delivery services [1, 20,21]. This advancement, however, is exacerbating the issues and risks connected with big data. Daily, billions of data are being generated from these applications and devices. However, not a lot is being translated into meaningful data.

There are lots of advancements in AI in the healthcare industry, the most common application is critical health diagnosis [13, 14, 17, 18], robots that perform or assist in surgical operations [5], and textual features observation for physological status detection [19,22]. This paper will attempt to shine a light on other not-sowell-known applications of mobile health data. The data collected with mobile health applications with the aid of sensors will be to make making decisions and predictions based on the findings of the model. The models used in this research can be used in the sports industry. This study explores the utilization of mobile applications on smartphones and tablets to gather information from people working in different sports and fitness environments, such as coaching effectiveness, strength and conditioning, and fitness training. After further exploration, we believe m-health can be implemented at a commercial level to determine which player will be fit to play games based on the data collected with sensors before a match. It will assist coaches, physicians, or any entity to make decisions using the model we have proposed.

## II. RELATED WORK

Yu et al. [2], the journal focused on how to use blockchain technology to collect health data to advance the sports industry. The journal believed blockchain being decentralized and secure makes it ideal for the sports industry. The collected data wouldn't be altered to make the athlete compete if found to be unfit. This problem might be noticed in the IoT system that is being currently used to store m-health data. They emphasized how the athlete's personal condition can be better understood with the use of a better and more efficient sports health data gathering system and in turn help players better target their training. Various limitations in the sports sector can be remedied by using blockchain technology owing to blockchain technology's independence as well as transparency. Various human health and sports indicators are assessed at any time and in any location to give a scientific foundation for human health evaluation. Blockchain technology has improved the authenticity of the sports industry sector and structural adjustment policies for creating sports health programs, which is extremely beneficial for the advancement of the sports industry. The methodology they used was a scientific formula evaluation to be more specific they used an accuracy index and a formula requested data recovery and carried out a study of demand to assess the physical fitness of an athlete. They calculated the accuracy index using the precision and recall formula. They also observed multi-mode node sensors giving better accuracy than the single-node ones.

Salsavil et al. [6] proposed myocardial disease prediction by applying a classification model. In this paper they try to also predict human fitness in terms of collecting blood pressure data and activity data like running, cycling, and standing at a particular time. Then they analyze the data and applied a machine learning algorithm to predict myocardial disease in order to give a decision for a particular person whether he is fit or not physically. The model they used is clustering and the accuracy they got was 84.66%.

Maszczyk et al. [8] In this model They are using a mathematical model established in this research, depending on the neural network, which attempted generalization and forecasting, neural networks were applied to improve the recruitment and selection procedures for javelin throwing. for generalization and prediction, which supported the application of neural networks in the optimization of the recruiting and determining system for javelin throwing. According to the findings of several statistical tests, the generated linear model for the examined group of teenage javelin throwers was unable to accurately depict the connection between each of the independent factors and the dependent variable

Kerdjidj et al. [10] proposed a reliable automatic fall detection system that is also suitable for detecting various daily activities (ADL) to predict human fitness. In the processes of collection and classification, just the accelerometer's data are used initially after that combination of the readings from the accelerometer and gyroscope is considered. The two configurations are contrasted, and it is demonstrated that the system created by combining CS capabilities may attain accuracy levels of up to 99.8%.

Cheng et al. [11], the authors looked at the connection between physical activity and weight status. On a bigger dataset, the effectiveness of various machine learning algorithms was assessed. The study's goal was to discover a connection between exercise and obesity in order to get fitness physically. However, there were no sensory data utilized to categorize or record physical activity.

Rossi et al. [12], the author proposed this narrative review to give instructions on how to train, validate, and test machine learning models that forecast sporting events. This narrative review's key contribution is to draw attention to any strengths and weaknesses that might exist at each stage of model construction, including training, validation, testing, and interpretation, in order to reduce misinformation that might lead to inaccurate results. The sample injuries forecaster in this paper illustrates all the characteristics that can be utilized to predict injuries, as well as all the various time series analysis pre-processing methods, how to perfectly The importance of describing the decision-making process on the whiteboard as well as splitting the dataset to train and validate the forecasting analytics.

Muazu et al. [15], in this article, mainly identified and forecasted high and low prospective archers using a variant of k-NN algorithms and logistic regression. One-end archery shot score test was completed by 50 young archers from different archery programs, with a mean age and standard deviation of (17.0 + 0.56 years). The handgrip, vertical leap, standing wide jump, static balance, upper- and core-muscle strength tests, as well as other common fitness assessments, were performed. The study showed that the vertical leap and core muscle strength were statistically significant. The study showed that the vertical leap and core muscle strength were statistically significant. The archers were grouped according to the significant variables found using hierarchical agglomerative cluster analysis (HACA). Based on the significant performance characteristics, different k-NN model variants, such as fine, medium, coarse, cosine, cubic, and weighted functions, as well as logistic regression, were trained. Archers were divided into high potential archers (HPA) and low potential archers by the HACA (LPA). When it came to the classification of the examined indicators, All the other evaluated models were outperformed by the weighted k-NN, which had a prediction accuracy for the HPA and LPA of 82.5 4.75%. Further research into the classifiers' performance using new data further revealed the effectiveness of the weighted K-NN model. These results may help coaches and sports management identify highpotential archers by combining a few of the physical fitness performance markers they have already established, saving money, time, and effort on a talent discovery program.

Oytun et al. [16] The Author proposed forecasting specific athletic performance categories displayed by female handball players, and various machine learning models were evaluated in this study. The superior model was then used to identify the major aspects influencing anticipated performances and effectiveness of women's handball players was predicted using linear regression, decision trees, support vector regression, radial-basis function neural networks, backpropagation neural networks, and long short-term memory neural network models in handball agility tests, 10- and 20-meter sprints, hands-onhips and hands-free dynamic balance leaps, a 20-meter shuttle run test, and 10- and 20-meter sprints. For each machine learning model, 118 instances of training patterns, 23 properties and measurements of the attribute, and a total of 23 attributes were recorded. According to the findings, the radial-basis method neural network performed better than the other models and had R 2 values between 0.86 and 0.97 for its ability to predict the various forms of athletic performance. Finally, by retraining the improved model, key factors influencing expected performance were identified. The findings from this study, one of the first to use machine learning in sports sciences to analyze handball players, are promising for future research.

III. METHODOLOGY

In this paper, we will analyze the body motion and vital sign recordings of ten volunteers while they perform different activities. Sensors were placed in different parts of their body. Our study used the collected data to analyze and monitor the subject's health and make predictions from patterns observed by using several machine learning algorithms. This paper will also address the research issues seen in Mobile Health data and the future advancements it will bring. Data collection is the initial stage of the technique in this study, which is then followed by a plan for preprocessing the dataset. The dataset is used for training and testing a variety of classifiers, including XGBoost, Logistic Regression, Random Forest, Decision Tree, and Naive Bayes. By analyzing the machine learning model one particular team can make the decision on which player is a good fit or an immediate game. To find the most accurate Athletics fitness prediction method, the results are calculated and applied.
Figure 1 shows the overall layout of the suggested study.

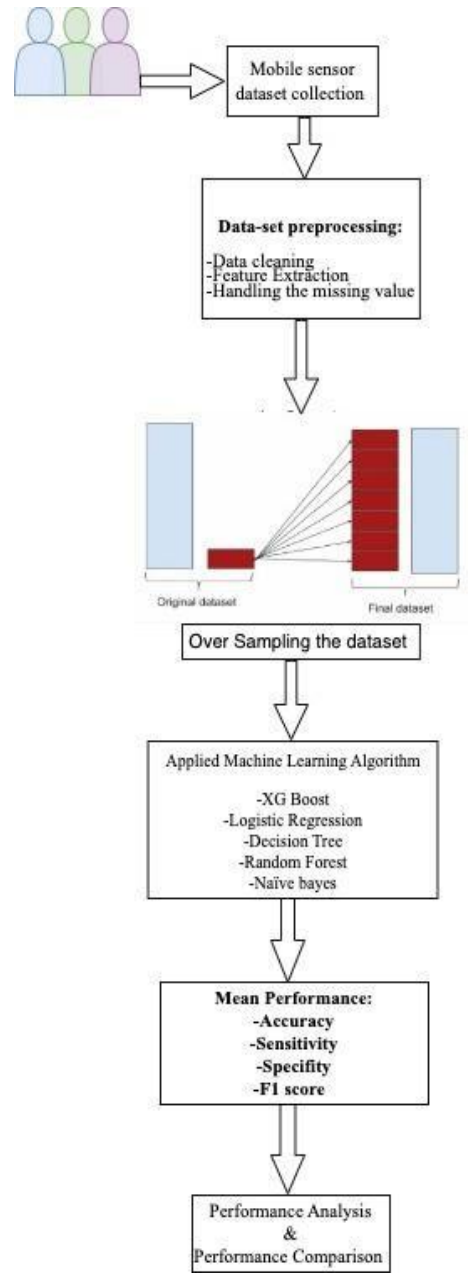

Figure 1: The overview of the study

### A. *Dataset collection & Data preprocessing:*

The first step for this paper was collecting the dataset This dataset has 1048576 instances, and 13 attributes, whereas 12 attributes are predictive or independent variables (physical activity set), and 1 class attribute is considered as fitness activity or dependent variable. The dataset contained body movements and vital signs for ten volunteers with various profiles while completing 12 physical activities such as Standing still (1 min), Sitting and relaxing (1 min), Lying down (1 min), Walking (1 min), Climbing stairs (1 min), Waist bends forward (20x), Frontal elevation of arms (20x), Knees bending (crouching) (20x), Cycling (1 min), Jogging (1 min), Running (1 min), Jump front & back (20x) [4]. The recordings were taken with wearable sensors called Shimmer2 [BUR10]. We consider these 12 physical activities because of their direct relationship with some health conditions such as heart rate, Blood pressure, breathing rate, and mental condition, etc which generate 48576 amounts of physical data for the activity already above mentioned. By getting those health data we can take the decision which algorithm might give the best result to take the decision whether one player is fit or not for the next game. The sensors were fastened using elastic this subject was bound around the left ankle, right wrist, and chest. By application of several sensors allows us to better capture the movement that different body components experience, such as the direction of the magnetic field, the frequency of turn, and acceleration. The sensor on the chest also gives 2-lead ECG values [5]. This data can be utilized for basic cardiac monitoring, testing for various arrhythmias, and examining the impact of exercise on the ECG, among other things. The sampling rate for all sensing modalities is 50 Hz, which is considered enough for capturing human activity [4]. A video camera was used to capture each session. The activities were gathered in an out-of-lab setting with no restrictions on how they had to be carried out. The activity set the volunteers performed consisted of a minute of standing still, sitting and relaxing, lying down, walking, climbing stairs, cycling, jogging, and running. They also had to perform 20 repetitions of frontal elevation, waist bends forward, and knee bending(crouching). We are illustrating the raw data, filtered data, and changing the label 'activity' exists a gap of ~1000 samples (~20sec in sample rate 50Hz) graph below which gives the audience an idea of how data is actually formed through the sensor.

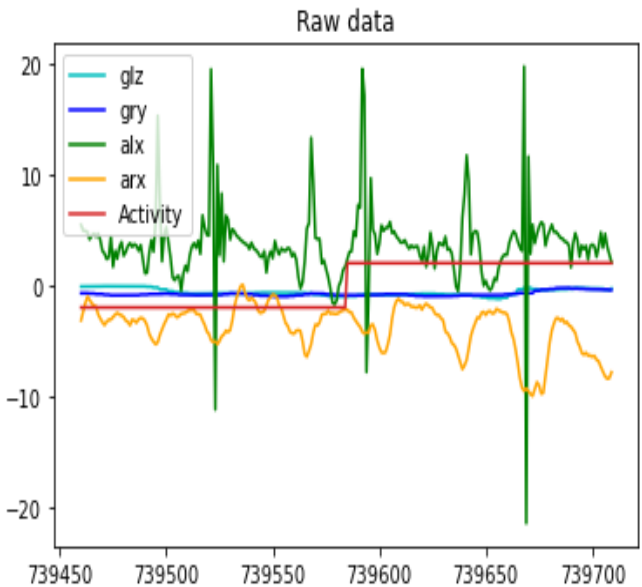

Figure 2: Raw data

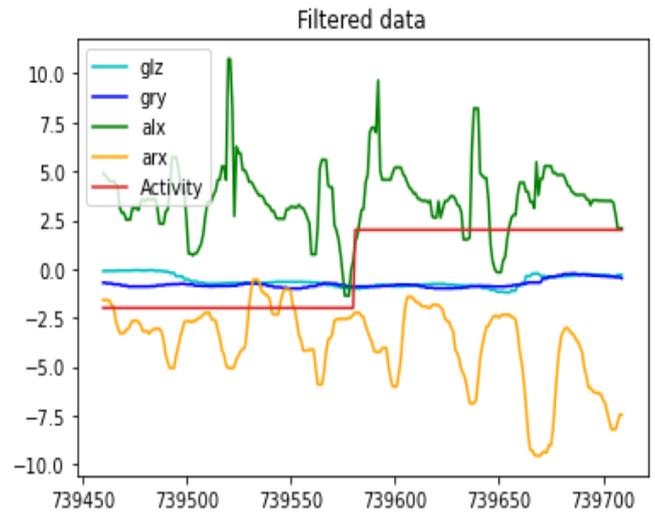

Figure 3: Filtered data

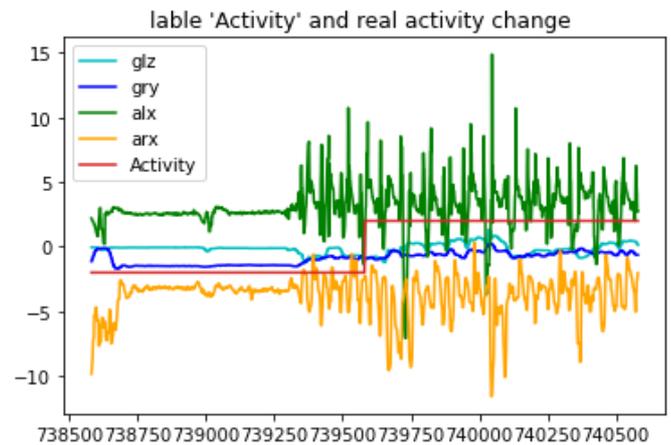

Figure 4: Label activity and real activity change

B. *Validation process:*

choosing the proper validation technique for specific Datasets is essential. The hold-out validation method is the most effective strategy [8]. we are training 70% of the dataset and testing 30% of it, we applied a hold-out validation technique to get good results. Furthermore, we measured the accuracy, sensitivity, specificity, and F1- Score by the implied confusion Metrix. A thorough examination is provided in the visualization and display of the performance indicators bar graphs. We have also summarized the process in a flowchart a breakdown of the study project's components.

IV. RESULTS AND DISCUSSION

In this paper, Table 1 depicts the evaluation of multiple machine learning models' performance used to predict physical fitness for athletics, including XGBoost, Decision Tree, Logistic Regression, Random Forest, and Naive Bayes. We used different types of performance metrics to evaluate our models such as accuracy, sensitivity, specificity, and F1 score. One of the most important performance metrics to evaluate how accurately the machine learning model performs is accuracy. Although logistic regression has the lowest accuracy of all the models in our chosen model (57.07%), XGBoost demonstrated superior accuracy in comparison to other models with a 95.92% accuracy rating.

We can determine the model performance by analyzing the sensitivity and specificity. Sensitivity and specificity for all of our selection model's performance show better results and are nearly identical, at 99% and 98%, respectively. The F1 score is another vital metric to evaluate models to analyze performance. In this research, most of the models show promising results except logistic regression (57.41%).

Table I. Results of different measures for different machine learning models predicting Athletics Fitness

| Models | Accuracy (%) | Sensitivity (%) | Specificity (%) | F-1 Score (%) |
|---|---|---|---|---|
| XGBoost | 95.92 | 99.50 | 99.50 | 99.66 |
| Decision Tree | 86.93 | 97.90 | 98.83 | 97.90 |
| Logistic Regression | 57.07 | 99 | 89.60 | 57.41 |
| Random Forest | 95.56 | 99.50 | 99.83 | 97.32 |
| Naïve Bayes | 92 | 99 | 97.41 | 97.32 |

After examining all of the results from each machine learning model, we determined that XGBoost performed better than other models in terms of accuracy (95.2%), sensitivity (99.5%), specificity (99.5%), and F1 score (99.5%). For the dataset we have chosen, XGBoost performs better than other models at predicting athletic physical fitness. Therefore, we may utilize XGBoost to estimate physical fitness and receive accurate and encouraging results.

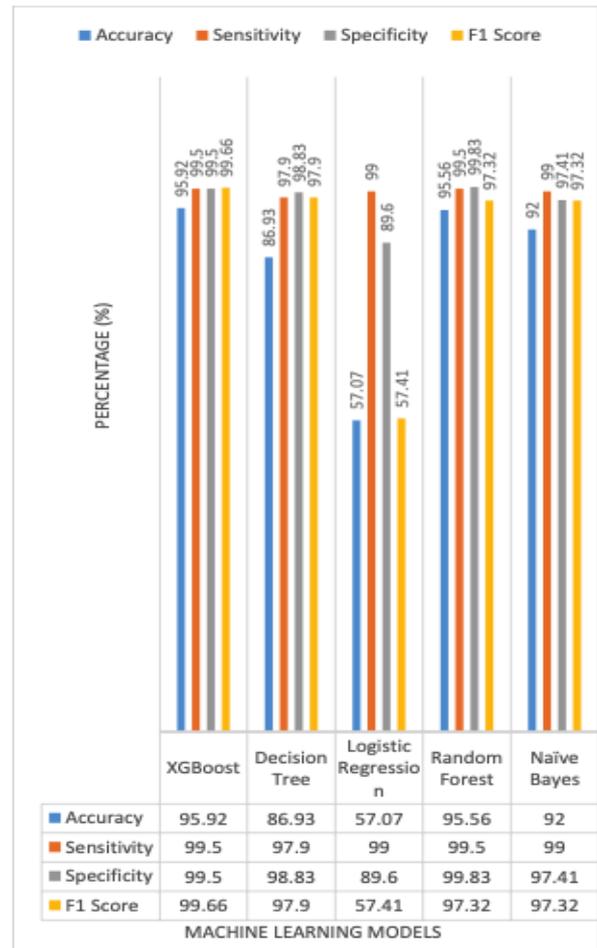

Figure 5: Performance analysis of machine learning models.

V. CONCLUSION AND FUTURE WORK

Many machine learning projects found XGBoost to outperform any other machine learning algorithm, this was also true within our study. We were able to get 95.2% accuracy, 99.5% in sensitivity, 99.5% in specificity, and 99.66% F-1 score using XGBoost. Our study was done with the data of 10 participants, if we wanted to scale this project XG Boost would still be the best algorithm to get better accuracy with great speed. It would not be right to assume the small sample we used to generalize for a bigger data set. More research needs to be done for this to be applied on a large scale. After big data issues are addressed and new technology like blockchain is adopted, m-health to predict human health and behavior analysis can come to fruition and tremendously help the industry progress. We might potentially focus on health fitness detection in the future. using different types of vocal features-based datasets by deep learning models.